\documentclass[fleqn,10pt]{wlscirep}
\usepackage[utf8]{inputenc}
\usepackage[T1]{fontenc}
\usepackage[sort&compress]{natbib}
\setcitestyle{super,comma,sort&compress,open={},close={}} 

\usepackage{soul}	
\soulregister\cite7  
\soulregister\citet7  
\soulregister\citep7  
\soulregister\ref7
\soulregister\textbf7

\soulregister\textit7

\usepackage{xcolor}
\usepackage{lineno}
\usepackage{multirow}
\usepackage{tcolorbox}

\title{Quantifying artificial intelligence through algorithmic generalization}

\author[1]{Takuya Ito}
\author[1]{Murray Campbell}
\author[1]{Lior Horesh}
\author[1]{Tim Klinger}
\author[1]{Parikshit Ram}
\affil[1]{Mathematics \& Theoretical Computer Science Department, T.J. Watson Research Center, IBM Research}
\affil[ ]{Contact: taku.ito1@gmail.com}

\begin{abstract}

The rapid development of artificial intelligence (AI) systems has created an urgent need for their scientific quantification.
While their fluency across a variety of domains is impressive, AI systems fall short on tests requiring algorithmic reasoning -- a glaring limitation given the necessity for interpretable and reliable technology.
Despite a surge of reasoning benchmarks emerging from the academic community, no theoretical framework exists to quantify algorithmic reasoning in AI systems.
Here, we adopt a framework from computational complexity theory to quantify algorithmic generalization using algebraic expressions: {\em algebraic circuit complexity}.
Algebraic circuit complexity theory -- the study of algebraic expressions as circuit models -- is a natural framework to study the complexity of algorithmic computation.
Algebraic circuit complexity enables the study of generalization by defining benchmarks in terms of the computational requirements to solve a problem.
Moreover, algebraic circuits are generic mathematical objects; an arbitrarily large number of samples can be generated for a specified circuit, making it an ideal experimental sandbox for the data-hungry models that are used today.
In this Perspective, we adopt tools from algebraic circuit complexity, apply them to formalize a science of algorithmic generalization, and address key challenges for its successful application to AI science.

\end{abstract}

\begin{document}

\flushbottom
\maketitle

\thispagestyle{empty}

\section*{Introduction}

The recent evolution of modern artificial intelligence (AI) systems and large language models (LLMs) has led to the speculation that these systems may reason \citep{bubeck_sparks_2023,wei_emergent_2022,webb_emergent_2023,deepseek-ai_deepseek-r1_2025}.
Yet due to challenge of evaluating large models trained on massive pretraining datasets \citep{kim_uncontrolled_2022}, it is difficult to evaluate whether such models are truly exhibiting algorithmic reasoning abilities, or whether they instead regurgitate plausible text from their pretraining data \citep{schaeffer_are_2023,wu_reasoning_2024}.
This ambiguity has led to a deluge of reasoning benchmarks \citep{lake_generalization_2018,hupkes_compositionality_2020,hudson_gqa_2019,yang_dataset_2018,ito_generalization_2024,johnson_clevr_2017,clark_boolq_2019,kim_cogs_2020,ruis_benchmark_2020,keysers_measuring_2020}.
Despite these efforts, objectively quantifying the complexity of reasoning problems is difficult; most of these experiments are ad hoc, and designed without a framework to quantify and verify the algorithmic complexity of reasoning problems.
However, approaches in computational complexity theory, a field within theoretical computer science, have made it possible to explicitly measure a problem's algorithmic difficulty, paving the way for generalization tests rooted in quantifiable measures of complexity.
In this Perspective, we bridge algorithmic reasoning with a decades-old branch of computational complexity -- \textit{circuit complexity theory} -- to provide a theoretical link to studying the complexity of algorithmic computation in modern AI systems.

Recently, there has been increased interest in studying modern AI models through arithmetic and compositional tasks \citep{kudo_deep_2023,mcleish_transformers_2024,zhou_what_2024,shen_positional_2024,saxton_analysing_2019,lee_teaching_2023,dziri_faith_2023,mccoy_embers_2023,zhou_transformers_2024,ito_generalization_2024,hupkes_compositionality_2020,sinha_survey_2024}.
Compositional tasks are problems rooted in a long history from the early 20th century\citep{moravcsik_thought_1975,carnap_meaning_1988} that are generated by recombining a basis set of atomic elements to form a variety of task combinations (for a review, see \citet{russin_frege_2024}).
(Arithmetic problems are compositional; they are composed of atomic elements (numbers and operators), and can be recomposed to generate novel expressions and problems.)
For modern AI systems, compositional tasks can serve as useful reasoning benchmarks since they require 1) abstraction, 2) logical and \textit{verifiable} application of rules or axioms, and 3) precise problem-solving and rigor.
Critically, these paradigms have provided reliable ways to elicit failure modes in transformer-based AI models for specific forms of compositional generalization.
For example, a number of studies have demonstrated the difficulty of ``length generalization'' -- generalizing to problems of longer length than seen during training \citep{kazemnejad_impact_2023,mcleish_transformers_2024,zhou_transformers_2024,shen_positional_2024}.
Other researchers have also introduced various notions (e.g., systematicity and productivity) in an effort to taxonomize different forms of compositionality \citep{hupkes_compositionality_2020,hupkes_taxonomy_2023,fodor_connectionism_1988,fodor_connectionism_1990,smolensky_constituent_1991}.
By contrast, the formalisms from circuit complexity theory provide a set of tools that can be applied to quantify \textit{algorithmic generalization} -- generalization over algorithms specified by circuits and their associated measures of complexity, such as space or time complexity.
(Structural properties of circuits correspond directly to algorithmic requirements to compute that problem.)
Moreover, formalizing problems through a circuit complexity framework provides a theoretically grounded framework for the increasingly popular, yet nascent empirical evaluations in AI systems that use arithmetic and compositional tasks \citep{kudo_deep_2023,mcleish_transformers_2024,deletang_neural_2022,zhou_teaching_2023,shen_positional_2024,saxton_analysing_2019,lee_teaching_2023,dziri_faith_2023,ruoss_randomized_2023,jelassi_length_2023,wang_exploring_2021,nogueira_investigating_2021}.
Though we focus on formalizing algebraic problems through the lens of circuit complexity (i.e., \textit{algebraic circuit complexity}) due to the widespread use of arithmetic problems to evaluate modern AI systems, the more general framework of circuits can be similarly extended to other algorithmic problems.

\begin{figure}[h!]
\centering
\includegraphics[width=6.93in]{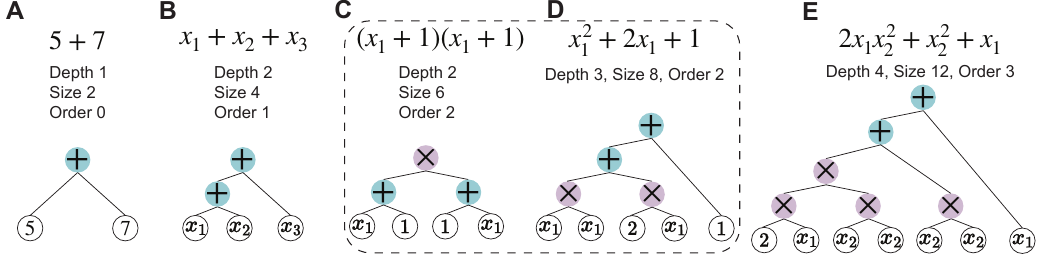}
\caption{
Examples of algebraic expressions represented as circuits.
A) A two operand addition circuit (input gates are sampled from a field $\mathbb{F}$).
B) A three operand addition circuit (input gates are sampled from the set of variables $x_i \in X$ rather than $\mathbb{F}$).
C, D) A mathematically equivalent pair of circuits, but represented as C) a factorized expression, and D) its a monomial expansion. 
Notably, despite their mathematical equivalence, the circuit representations are distinct.
E) A polynomial of depth 4 and size 12.
}
\label{fig:fig1}
\end{figure}

A large class of algorithmic problems can be studied with algebraic expressions \citep{shpilka_arithmetic_2010,burgisser_algebraic_2013}. 
Algebraic circuit complexity theory formalizes the evaluation of algebraic expressions through algorithms encoded as a computable circuits (i.e., directed acyclic graphs; Fig. \ref{fig:fig1}).
This formalization is well-established in computational complexity theory, the branch of theoretical computer science concerned with quantifying the difficulty of computational algorithms and the resources required to solve them \citep{arora_computational_2009}.
Importantly, formalizing computational problems in terms of circuits is the leading approach to empirically quantify their complexity.
Unlike other notions of complexity, such as Kolmogorov Complexity in algorithmic information theory (which is incomputable), notions developed in complexity theory for circuits are {\em explicitly computable} and determined by their shape and structure \citep{wyeth_circuit_2023}.
Thus, the tools of circuit complexity can formalize notions of generalization over algorithms by defining benchmarks in terms of their circuit properties.
Furthermore, algebraic circuits are generic mathematical objects; they can be represented from a variety of mathematical perspectives (geometry, topology, etc.), providing useful interpretations in other domains.
Algebraic circuits are therefore well-situated to investigate algorithmic generalization -- the problems are computable and verifiable, large datasets can be straightforwardly generated from circuit specifications, and new models can be developed that address specific failure modes within this framework.
In the ensuing sections, we provide a blueprint for the successful adoption of algebraic circuit complexity as a gateway into studying algorithmic generalization more broadly.
We introduce the core components of algebraic circuits, address how they can be leveraged to study algorithmic generalization, and discuss several key open theoretical and empirical challenges.

\section*{Algebraic circuits}

Algebraic circuit complexity studies algebraic expressions as computable circuit models. 
There has been a significant amount of recent machine learning research studying arithmetic generalization, a key algorithmic ability.
While important and insightful, most of those studies primarily focus on a restricted set of algebraic problems, which we will illustrate later \citep{dziri_faith_2023,kudo_deep_2023,mcleish_transformers_2024,shen_positional_2024,jelassi_length_2023}.
Below, we provide definitions of algebraic circuits that will place prior work within a broader mathematical framework.
Our goal is to provide the tools to quantify model generalization through circuit complexity, rather than ad hoc goals such as, for example, ``length generalization''.

\noindent {\bf Definition.} 
(For a more formal definition, see \citet{shpilka_arithmetic_2010} or \citet{burgisser_algebraic_2013}.)
An algebraic circuit $C$ represents a polynomial expression as a directed acyclic graph, comprised of gates $v$ and edges $e$. 
Input gates, $v_X$ and $v_{\mathbb{F}}$, are respectively defined as either variables (e.g., $X=\{x_1,...,x_n\}$) or elements in a field $\mathbb{F}$ (e.g., $\mathbb{R})$. 
Input gates have fan-in (in-degree) of 0.
All other gates are operators: a sum gate ($v_+$) or a product gate ($v_{\times}$).
In this Perspective, we restrict the fan-in of operator gates to 2, as is standard \citep{shpilka_arithmetic_2010}.
For our purposes, this ensures that there is the same number of gates in a circuit model with its corresponding representation as a string of tokens, which simplifies downstream analyses (e.g., the analysis in Fig. \ref{fig:fig6}).
Operator gates have a fan-out (out-degree) of either 1 or 0. 
If the fan-out of an operator gate is 0, then it is the output gate of that polynomial.

\noindent {\bf Properties.} 
An algebraic circuit $C$ has two main algorithmic complexity measures: size $s$ and depth $d$ (see Table \ref{table:table1} for a summary of all properties).
The size $s$ of a circuit refers to the number of edges $e$ in $C$, and corresponds to algorithmic space complexity.
The depth $d$ of a circuit refers to the longest path from an input gate to the output gate, and corresponds to algorithmic time complexity.
We can denote a {\em sub-circuit} $C_v$ of $C$, which computes the polynomial $f_v$ rooted at gate $v$.
Another important property of an algebraic circuit is its {\em degree} (i.e., the degree of a polynomial). 
The degree of a circuit (or a sub-circuit) can be computed by measuring the degree of the gate $v$, denoted $deg(v)$.
Elements in a field $\mathbb{F}$ are of degree 0, input variables $x \in X$ have degree 1, the degree of an sum gate ($+$) is determined by the degrees of its inputs $u$, $v$ such that $deg(u+v)=max(deg(u),deg(v))$, and the degree of a product gate ($\times$) is determined by $deg(u \cdot v)=deg(u) + deg(v)$.

\begin{table}[h!]
\centering
\begin{tabular}{c | c}
\cline{1-2}
{\bf Circuit property} & {\bf Description} \\
\hline
 Circuit $C$ & A directed acyclic graph that computes a polynomial. \\
\hline
 Size $s$ & The number of edges in $C$. \\
\hline
 Depth $d$ & The longest path from an input gate to an output gate. \\
\hline
 $deg(v)$ & The degree of a gate $v$. If $v$ is the output gate, $deg(v)$ is the degree of the polynomial. \\
\hline
 Gate $v_{\mathbb{F}}$ & An element of a field $\mathbb{F}$ with $deg(v_{\mathbb{F}})=0$, and fan-in 0. \\
\hline
 Gate $v_X$ & A variable $x_i \in X$ with $deg(v_X)=1$, and fan-in 0. \\
\hline
 Gate $v_+$ & A sum gate with $deg(v_+)=max(deg(u),deg(v))$ for inputs $u$ and $v$, and fan-in 2. \\
\hline
 Gate $v_{\times}$ & A product gate with $deg(v_{\times})=deg(u) + deg(v)$ for inputs $u$ and $v$, and fan-in 2.
\end{tabular}
\caption{Key circuit properties and their descriptions.}
\label{table:table1}
\end{table}

Figure \ref{fig:fig1} provides a few simple examples of algebraic circuits.
A circuit representation of an algebraic expression provides a concrete algorithm for {\em computing a polynomial}, with some circuits requiring more computation than others (e.g., determined by size or depth).
Furthermore, a circuit description can provide explicit differences in required computation for polynomials that are mathematically equivalent (e.g., see Fig. \ref{fig:fig1}C,D).
Such a formalization can be useful to study how different representations of equivalent algebraic expressions can lead to different levels of generalization, or how syntax relates to semantics in formal languages.
The recent studies in arithmetic generalization in AI models focus on the simplest circuit representations of algebraic problems (problems analogous to Fig. \ref{fig:fig1}A, but varying the magnitude of the field elements, e.g., train on small digit numbers, test on large digit numbers). 
The language of circuit complexity can provide more interesting and flexible ways to investigate generalization across circuit complexity classes.

\section*{Towards a science of generalization with algebraic circuits}

A recent study concluded that LLMs' behavior is a function of the problems they are trained to solve \citep{mccoy_embers_2023}.
It is difficult to identify what the ``problems'' are when the pretraining corpus is natural language text derived from the internet.
A theoretically coherent alternative to quantify AI systems is to select problems where 1) complexity is quantifiable, and 2) arbitrarily large datasets can be systematically generated.
Algebraic circuits satisfy both of these constraints.

\begin{figure}[ht]
\centering
\includegraphics[width=6.93in]{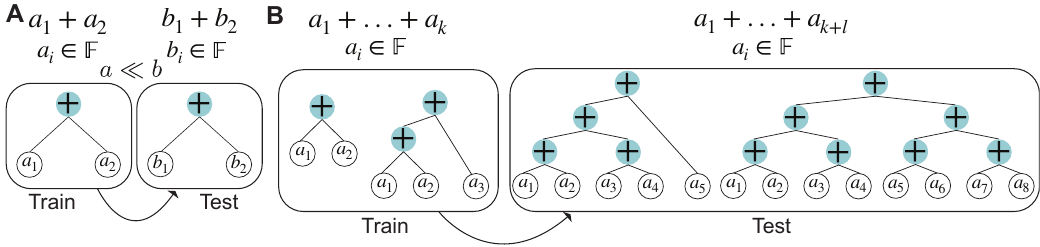}
\caption{
Commonly-used AI evaluations for ``length generalization'' with arithmetic tasks.
A) The predominant form of ``length generalization'' in transformers is evaluating performance on addition or multiplication problems with larger integers than seen during training \citep{shen_positional_2024,jelassi_length_2023,mcleish_transformers_2024,zhou_transformers_2024,lee_teaching_2023,zhou_what_2024}.
This would be conceptually equivalent to a situation in which the circuit size and depth are fixed, but the sampling of input gates (i.e., field elements) differs across training and testing sets. 
(However, see discussion below in \textit{Open theoretical and empirical challenges} on the nuances of computing long addition.)
The notion of length generalization is specific to the context of transformers, given that larger digits require a larger context window.
B) Another form of ``length generalization'' studied in the literature is to generalize to arithmetic problems with more operands (variables) than seen in the training set \citep{deletang_neural_2022}.
From a circuit complexity perspective, these two approaches are distinct problems.
}
\label{fig:fig2}
\end{figure}

Though recent papers have demonstrated an increased interest in using arithmetic tasks to quantify generalization, preliminary approaches have been theoretically limited.
As alluded to above, many of the recent papers evaluating ``length generalization'' in arithmetic tasks train on two variable addition (or multiplication), and test on the same circuit class but sample field elements that are larger in magnitude (Fig. \ref{fig:fig2}A) \citep{zhou_what_2024,shen_positional_2024,nogueira_investigating_2021,mcleish_transformers_2024,zhou_transformers_2024,lee_teaching_2023,jelassi_length_2023}.
Others focus on a more complex form of generalization, i.e., addition or multiplication on problems with a greater number of variables, which is equivalent to increasing the circuit size and depth (Fig. \ref{fig:fig2}B) (e.g., modular arithmetic task; \citet{deletang_neural_2022}). 
While useful and informative, both these tests of generalization scratch the surface of generalization metrics that can be devised with algebraic circuits.

\subsection*{The importance of learning composable functions}
We first define algorithmic generalization (Box 1), and consider the importance of learning algorithmically.
The ability to generalize algorithmically is one of the most challenging problems in AI, and is a requirement for robust reasoning and planning.
Like algebraic circuits, several prior papers formulated algorithmic tasks as computing a path through a directed graph or circuit \citep{dziri_faith_2023,yang_dataset_2018,johnson_clevr_2017,ram_what_2024}.
Importantly, learning an algorithmic task requires 1) decomposing and abstracting the individual functions (e.g., nodes) of a circuit, and understanding how they can be recomposed via edges to produce novel circuits (e.g., function composition).
Algebra is a natural language to study algorithmic generalization as it can be encoded as a circuit, and requires both abstraction and function composition.
Moreover, in contrast to other compositional approaches (e.g., regular or context-free grammars; \citet{deletang_neural_2022,chomsky_three_1956}), algebraic problems encompass an infinite vocabulary (e.g., $\mathbb{R}$) with an expressive grammar determined by its axioms, making it well-suited for machine learning algorithms that require many training samples.
In the following section, we propose an algorithmic generalization framework in terms of algebraic circuits.

\begin{tcolorbox}[colback=blue!5!white, colframe=blue!75!black, title=Box 1. A formal sketch of algorithmic generalization.]

Let $B$ be a basis set of elements.
Let $U_1 = B$ and $U_{i+1} =\{u \mid (u_1 , u_2 \in U_{\leq i}) \text{ \& } (u = u_1 \circ u_2) \}$, where $\circ$ stands for any computable composition operator on $U_i$ (akin to production rules in context-free grammars or operations in a field).
Then, we define some universe set 
$$U = \bigcup_{i \in \mathbb{N}} U_i $$
We define a function (task) $T$ on $U$ such that $T: U \rightarrow R$ (see below for tasks $T$ for algebraic circuits).
$T$ is a compositional function, where $\forall u_1, u_2 \in U$, $T(u_1 \circ u_2) = T(u_1) \ast T(u_2)$, and $\ast$ stands for any computable composition on $R$.
\newline

Let $D_{train}$ and $D_{test}$ be distributions over $U$, and $supp(D_{train} | B)$ and $supp(D_{test} | B)$ to be the support of the basis set elements in $D_{train}$ and $D_{test}$, respectively.
We restrict $D_{test}$ to be a distribution in which $supp(D_{test}|B) \subseteq supp(D_{train}|B)$.
Additionally, if a composable function $\circ$ is represented in $D_{test}$ (i.e., $u \in D_{test}$ and $u = b_1 \circ b_2$), then we require that $\circ$ is also represented in $D_{train}$.
We include these as requirements for algorithmic generalization; it would be challenging, if not impossible, to generalize to samples in $D_{test}$ if not all basis elements and composition operators were provided in $D_{train}$.
\newline

We define algorithmic generalization as the evaluation of a learned model $M$ (e.g., a neural network) where
\begin{equation}
    Pr_{x\sim D_{test}}[M(x)=T(x)] > 1 - \epsilon
    \label{eq1}
\end{equation}
Here, $M$ is only optimized from samples $x \sim D_{train}$, and $\epsilon$ is the evaluation error of $M$ on $x \sim D_{test}$. 
{\em A strong form of algorithmic generalization over algebraic circuits would be to satisfy Equation \ref{eq1} for all valid partitions of $D_{train}$ and $D_{test}$ (with the requisite support of basis sets and composition functions) in $U$ for uniformly small values of $\epsilon$.}
\newline

For algebraic circuit problems, $T: U \rightarrow R$ can be a function that maps a circuit to: 1) a field element, e.g., $r \in \mathbb{R}$ (in the case of circuit evaluation; Fig. \ref{fig:fig1}); 2) $\{0, 1\}$, in the case of a classification task, such as polynomial identity testing (Fig \ref{fig:fig3}A); 3) another circuit $C$ (i.e, in the case of polynomial expansion or factorization; Fig \ref{fig:fig3}B). 

\label{box1}
\end{tcolorbox}

\subsection*{Circuit divergence as a metric of generalization}

\begin{figure}[ht]
\centering
\includegraphics[width=6.93in]{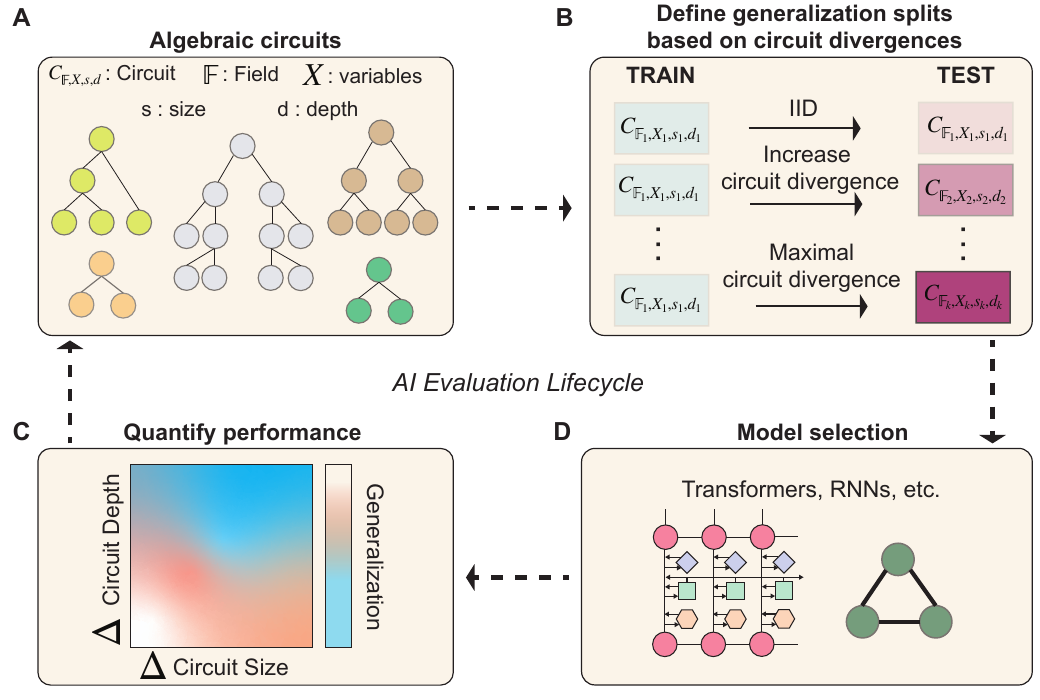}
\caption{
Algorithmic capabilities of modern AI systems and architectures can be studied with algebraic circuits. 
A) A set of problems can be identified (and sampled) from algebraic circuits.
B) Given a family of circuits, we can identify circuit divergences -- the divergence of different circuit properties, like size and depth -- to design train and test datasets to evaluate the algorithmic generalization capabilities of a model.
C,D) We can evaluate a model (or set of models) across these circuit splits, and quantify their performance according to circuit divergences.
This AI evaluation lifecycle allows us to iterate and refine hypotheses regarding the degree to which a model can generalize to a class of algorithms.
(Note that circuit divergences can be measured beyond depth and size.)
}
\label{fig:fig3}
\end{figure}

The language of algebraic circuits provides quantitative metrics to formalize generalization in terms of algorithmic complexity, since circuit structure directly corresponds to algorithmic properties.
In particular, generalization of AI systems can be quantified in terms of {\em circuit divergence} -- the divergence of circuit parameters between algebraic circuits employed during training and testing (Fig. \ref{fig:fig3}).
This emphasis on characterizing the algorithmic properties of problems complements prior work focused on establishing metrics of compositionality based on the degree of subgraph overlap in training and testing sets.
One particularly relevant metric is \textit{maximum compound divergence} (MCD) introduced by \citet{keysers_measuring_2020}, which measures the divergence of the frequency distribution of compositional subgraphs between the training and testing sets. 
While \citet{keysers_measuring_2020} demonstrated the empirical relevance of MCD on generalization performance, here we emphasize the importance of designing training and testing partitions according to the distribution of \textit{algorithmic complexity} (i.e., circuit) properties.
This contrasts with MCD, which focused on providing a single summary statistic that, though empirically useful, does not distinguish between algorithmic properties.
Specifically, designing benchmarks by manipulating circuit size (vs. depth) enables the characterization of generalization over algorithmic size (vs. time) complexity.
This in turn can help practitioners isolate how specific architectural components of models map onto algorithmic capabilities: for example, circuit depth of a problem typically relates to the number of layers of a transformer, while circuit size relates to the degree of parallelization, such as context length size for transformers \citep{strobl_what_2024}.)

Given a model $M$, we seek to characterize its generalization performance $M(C_{test} | \mathcal{C})$, where $C_{test}$ is a test circuit, and $\mathcal{C}=\{C_1, ..., C_k\}$ is a family of training circuits.
We define circuit divergence as the difference between quantifiable circuit properties between train and test distributions. 
Here, we emphasize five important circuit properties to measure divergences: size $s$, depth $d$, the polynomial degree, the sampling of a field $\mathbb{F}$, and the number of variables $\lvert  X \rvert$.
By quantifying the properties of circuits in both the training and testing sets, circuit divergence can be explicitly measured along these dimensions.
In the following sections, we provide several examples where tests of generalization can be constructed through the manipulation of circuit divergence.

\subsection*{Generalization benchmarks}

In this section, we illustrate the flexibility of algebraic circuits in designing meaningful AI benchmarks.
To demonstrate the generality of algebraic circuits, we begin by providing benchmarks that are analogous to popular tests of compositional generalization: systematicity and productivity \citep{hupkes_compositionality_2020}.
We subsequently introduce novel, more challenging problems that can be used to evaluate more general types of algorithmic computation and abstraction.
Finally, we demonstrate an approach to reformulate tests of algorithmic generalization for pretrained LLMs.

\subsubsection*{Compositional generalization with algebraic circuits}

There has been recent interest in using compositional paradigms to systematically study the generalization capabilities of machine learning models \citep{hupkes_compositionality_2020,lake_generalization_2018,lake_human-like_2023,ruis_improving_2022,dziri_faith_2023,ontanon_making_2022,csordas_devil_2021,ito_generalization_2024,klinger_compositional_2023,poggio_compositional_2024,ram_what_2024}.
Here we demonstrate direct links to common forms of compositional generalization using algebraic circuits.

\begin{figure}[h!]
\centering
\includegraphics[width=6.93in]{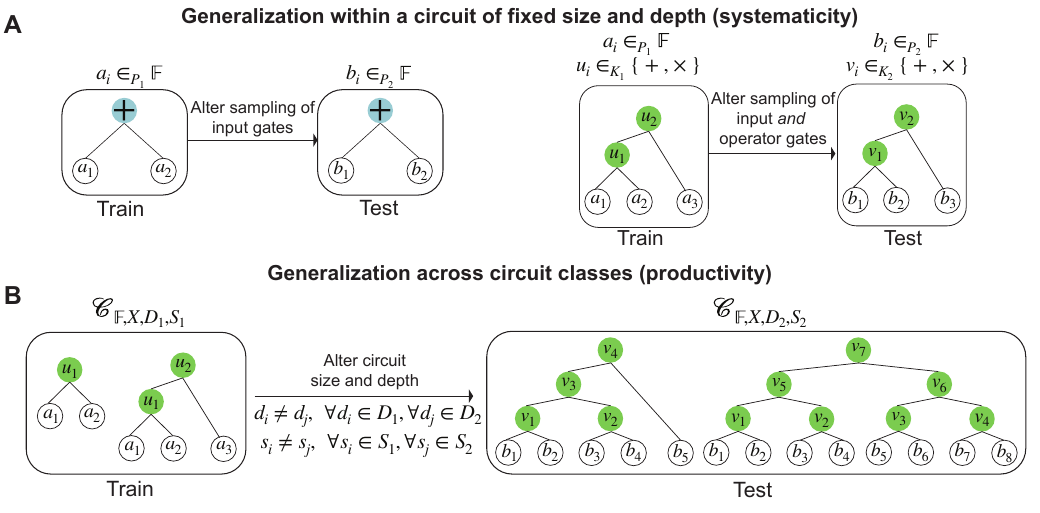}
\caption{
Analogs of common compositional generalization benchmarks in terms of algebraic circuits.
A) A simpler form of generalization within an algebraic circuit is generalizing to circuits of the same structure (size and depth), but with a novel combination of gates.
The analog of this in compositionality is commonly-referred to as {\em systematic compositional generalization} \citep{hupkes_compositionality_2020}.
(Learning over circuits of fixed structure can also be viewed as learning a regression model \citep{zhang_class_2022}.)
On the left, we illustrate an example of a model that is trained on a restricted family of circuits where the input gates are sampled from a field $\mathbb{F}$ with a sampling function $P_1$.
At test time, the model is required to generalize to circuits of the same circuit class (in terms of size and depth), but where the input gates are sampled using a different sampler $P_2$.
On the right, both input and operator gates are chosen with separate samplers across train and test circuits, resulting in a more difficult test of systematic compositionality.
B) Productive compositional generalization considers partitions of the training and testing sets across circuit sizes and depths. 
The analog of this in compositionality is commonly-referred to as {\em productive compositional generalization}.
Given a family of circuits $\mathcal{C}_{\mathbb{F},X,D,S}$, where $D$ and $S$ denote a set of depths $d_i$ and size $s_i$, the primary experimental manipulation is to construct a training set $\mathcal{C}_{\mathbb{F},X,D_1,S_1}$ and testing set $\mathcal{C}_{\mathbb{F},X,D_2,S_2}$ such that there exists no overlap of a specific circuit class $C_{\mathbb{F},X,d_i,s_i}$ between the training and testing sets.
}
\label{fig:fig4}
\end{figure}

\noindent {\bf Systematic compositional generalization and regression}. 
Systematic compositional generalization refers to the ability to recombine known basis elements into novel combinations of fixed sequence size (Fig \ref{fig:fig4}A).
This means that test sets of systematicity are limited to novel combinations of the same length as seen during training \citep{lake_generalization_2018,hupkes_compositionality_2020}.
The analog in algebraic circuits is to 1) sample a family of circuits $\mathcal{C}$ over a field $\mathbb{F}$ of a fixed size and depth, and 2) to generalize to circuits of the same size and depth, but differentially sampling input gates and/or operators.
This can be implemented by choosing different samplers -- $P_1$ and $P_2$ -- that differentially sample gates (Fig \ref{fig:fig4}A).
For example, the training set of circuits could be constructed with input gates $a_i \in_{P_1} \mathbb{F}$, and the testing set of circuits could be constructed with input gates $b_i \in_{P_2} \mathbb{F}$.
(Note that when samplers preferentially choose $b_i \gg a_i$, this is analogous to the common test of ``length generalization''; Fig. \ref{fig:fig2}A.)

In real world data, distributional properties of the training distribution can often bias AI (or even human) learners towards learning memorized short cuts, rather than learning algorithmic or syntactic strategies that enable robust generalization\citep{wu_reasoning_2024}.
However, given the infinite vocabulary (i.e., field elements) and the ability to sample circuits with well-defined algorithmic properties, algebraic circuit complexity provides a comprehensive experimental sandbox for designing and sampling from diverse data distributions.
Through this sandbox, practitioners can identify specific properties of train/test distributions that can either introduce a distributional bias/confound to be assessed, or ameliorate these biases by counterbalancing the distribution.
Interestingly, we note that successful systematic compositional generalization on input gates sampled from different distributions amounts to learning a distributionally robust regression model (e.g., \citet{zhang_class_2022,ghosh_efficient_2021}).
This is because a circuit of specific size and depth (e.g., as shown in Fig. \ref{fig:fig4}A) expresses a specific polynomial. 
Generalization over a fixed circuit with a distribution shift of its input gates (e.g., field elements but not operator gates) would demonstrate successful out-of-distribution generalization. 

Prior notions of systematic compositional generalization have specified ``weak'' and ``strong'' forms of systematicity\citep{hadley_systematicity_1994}.
``Strong'' systematicity refers to the ability to generalize to tokens (e.g., operands) in novel syntactic positions. 
(For example, if a model is exclusively trained on expressions of the form $a+b$ and $a+c$, where $a$ is always the first token, a model that generalizes to $b+a$ where $a$ is the last token would exhibit strong systematic generalization.)
``Weak'' systematicity refers to the ability to exclusively generalize to novel expressions in which $a$ is always in the same syntactic position (e.g., always the first token).
While this distinction does not alter the semantics for commutative algebras (as discussed in this paper), this distinction is important for many formal languages in which permuting syntactic ordering produces novel semantics, such as in many context-free grammars and non-commutative algebras.
(In these cases, circuits can be encoded as ordered DAGs, in which the order of parent/input gates is preserved.)

\noindent {\bf Productive compositional generalization.} Productive compositional generalization refers to the ability to generalize to sequences of longer length.
In the context of algebraic circuits, while systematicity focuses on keeping circuit size and depth fixed while manipulating gates, productive generalization focuses on manipulating circuit size and depth (Fig \ref{fig:fig4}B).
Thus, evaluating systematic and productive generalization can be studied together; measures of generalization can be quantified in a continuous manner by varying circuit parameters, such as gate samplers (systematicity), size (productivity), and depth (productivity). 
Understanding how each of these properties interact across training and testing sets will provide a comprehensive quantification of generalization over distinct algorithmic properties.

Since algebraic circuits are circuit representations of algebraic expressions, one can ask more generic questions about algebraic polynomials. 
For example: Given a class of polynomials as a training dataset, what other class of polynomials will this model be able to compute? 
Such a question goes beyond asking whether a model can systematically or productively generalize.
Instead, it addresses a basic question that can leverage other rich mathematical subfields (e.g., geometry, topology) to quantify algorithmic generalization.
Algebraic circuits provide a flexible framework to formalize problem complexity beyond existing paradigms in compositional generalization.

\subsubsection*{Classification tasks: Polynomial identity testing}

Prior work studying arithmetic abilities in transformer models have typically focused on computing simple expressions with field elements ($a \in \mathbb{F}$; e.g., $5 + 7 = ?$), rather than abstract variables ($x \in X$; e.g., $2x_1 + x_2 + 7$).
Including variables in an algebraic expression increases the polynomial degree ($d \geq 1$), thereby increasing its complexity and the need for abstractions.
One approach to evaluating algebraic circuits with variables is through {\em polynomial identity testing}, an active area of research in computational complexity theory and computational algebra.
Polynomial identity testing evaluates whether two polynomials are equivalent (i.e., $P_1(x_1,...,x_n) \equiv P_2(x_1,...,x_n)$). 
Posed another way, one can ask whether $P_1(x_1,...,x_n) - P_2(x_1,...,x_n) \equiv 0$.

\begin{figure}[h!]
\centering
\includegraphics[width=6.93in]{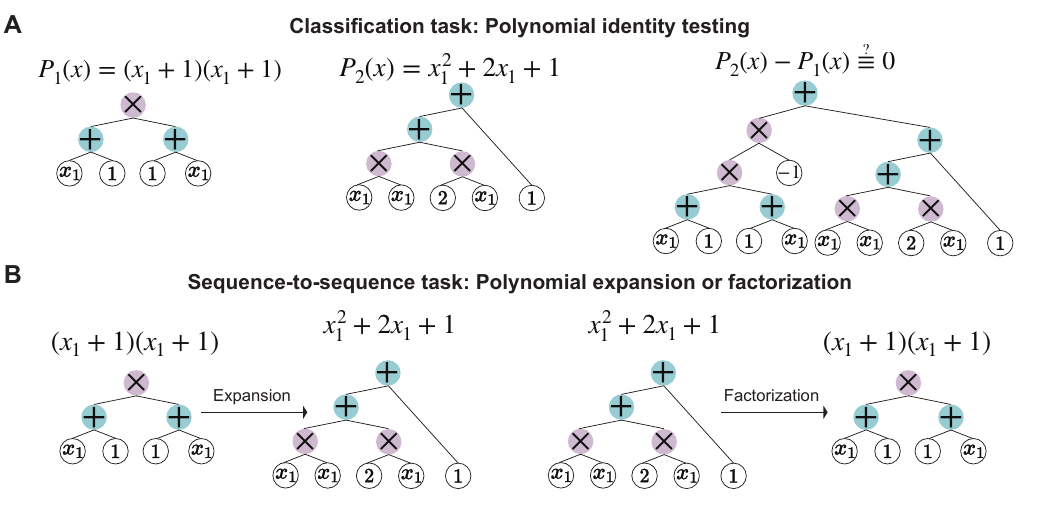}
\caption{
Algebraic problems as machine learning challenges.
Prior work that leverages arithmetic problems for machine learning studies are typically limited to evaluating expressions with field elements.
We introduce problems that can be evaluated with abstract variables.
A) Polynomial identity testing as a machine learning classification task. Polynomial identity testing, an important and active area of research in computational algebra, evaluates whether two polynomials are equivalent.
We illustrate two different polynomial expressions, $P_1(x)$ and $P_2(x)$ with distinct circuit representations, yet are mathematically equivalent.
To reformulate this as a classification task for machine learning studies, one can ask if $P_2(x) - P_1(x) \equiv 0$ (right).
B) Polynomial expansion and/or factorization as a transduction (sequence-to-sequence) task. 
Common sequence-to-sequence tasks in linguistics ask whether a model can expand a string using a set of rules or a grammar, such as SCAN or PCFG \citep{lake_generalization_2018,hupkes_compositionality_2020}.
In algebraic circuits, this is analogous to expanding a polynomial in a factorized representation (left).
Given the 1-1 correspondence of polynomials, an additional approach is to take a polynomial in its expanded form (sum of monomials), and generate the factorized representation (right).
This provides the ability to evaluate whether an AI system can expand an encoding (expansion), or compress an encoding (factorization).
}
\label{fig:fig5}
\end{figure}

This problem can be naturally formulated as a binary classification task for AI models.
Importantly, studying the extent of algebraic generalization by assessing train/test circuit divergences of polynomial identity problems can be rigorously quantified, since each identity problem can be encoded as a circuit (Fig. \ref{fig:fig5}A).
The complexity of a polynomial identity testing problem can be scaled up by increasing the size, depth, and/or degree of the test problems.
Moreover, machine learning practitioners can leverage existing symbolic programs and solvers to systematically generate large datasets (e.g., SymPy; \citet{meurer_sympy_2017}).
Polynomial identity testing thus provides a unique opportunity to study AI generalization in terms of circuit complexity theory. 

\subsubsection*{Sequence-to-sequence tasks: Polynomial expansion or factorization}

A wide application of generative models is in sequence-to-sequence transduction tasks.
One particularly impactful use case is the development of AI models for code. 
AI models for code take in code as inputs (e.g., COBOL), and generate a translation of that code (e.g., Java).
Despite its potential importance for modernizing many existing codebases, there is significant skepticism as to whether generative models trained with next-token prediction can reliably generate accurate code translations.
This is due to the fact that programming languages are not dictated by autoregressive processes, and instead governed by algorithmic rules.

Learning on algebraic circuits provides a straightforward framework to evaluate the ability of models to learn sequence-to-sequence tasks that are governed by algorithmic rules.
For example, the problem of expanding or factorizing a polynomial is a problem that is governed by the axiomatic rules of algebra (Fig. \ref{fig:fig5}B).
Like code translation, this task requires transforming one sequence into another while maintaining mathematical equivalence (or for code, semantic equivalence despite syntactic differences).
Importantly, there are explicit tools to describe the complexity with which the translation occurs in algebraic circuits. 
For example, a polynomial expansion transforms a factorized representation (i.e., shallow circuit) to a sum of monomials representation (i.e., deeper circuit).
In contrast, a polynomial factorization implements the inverse operation, which requires compressing a large circuit to a smaller circuit.
While a general interesting question is to ask whether models trained on one type of transformation can learn the other, this formalization has natural implications for understanding how to design AI models for code.
In particular, some programming languages may have lower-level syntax (e.g., COBOL) relative to other languages, such as Python or Java. 
In either case, understanding how lower-level and higher-level encodings of an expression (be they algebraic or programmatic) requires learning useful abstractions and the algorithms to translate them.
Thus, studying the conditions by which AI models can robustly parse and translate algebraic circuits can shed light on the best strategies to train AI models to translate programming languages.

\subsection*{Mechanistic interpretability with algebraic circuits}
\begin{figure}[h!]
\centering
\includegraphics{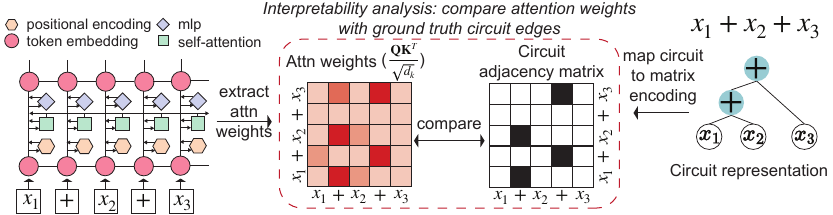}
\caption{Using an algebraic circuit's adjacency matrix as a ground truth comparison to interpret transformer attention representations.
The transformer's attention mechanism provides a useful way to peer into the representations of its input tokens (Left).
When the input is an algebraic expression (presented as a string of tokens), the attention matrix can be investigated to uncover the relationships between tokens (i.e., operators and operands).
On the other hand, an algebraic expression can always be mapped to a circuit encoding, which can be represented as the adjacency matrix of that circuit (Right).
This allows for the direct comparison between attention weights and the ground truth circuit representation (Middle).
This distance between the transformer's attention weights and the ground truth circuit adjacency matrix can also be used as a regularizer to encourage learning circuit algorithms via attention weights.
}
\label{fig:fig6}
\end{figure}

A major issue in assessing algorithmic generalization is the difficulty of interpreting what goes awry when they fail to generalize. 
This is partly due to the lack of interpretability of many benchmarks, which are often presented in natural language, and where verifiable algorithms (e.g., circuit diagrams or parse trees) do not exist \citep{clark_boolq_2019,hendrycks_measuring_2020}.
A more recent set of approaches in characterizing the interpretability of neural network representations rely on the design of carefully constructed tasks that have interpretable tasks, such as ground truth parse trees or circuit diagrams \citep{andreas_measuring_2018,dziri_faith_2023,deletang_neural_2022}.
Algebraic circuits similarly provide verifiable tasks \textit{and sub-tasks}.
While the input to an AI model might be a string representation of a polynomial, that polynomial's circuit encoding provides its ground truth encoding.
Such an encoding makes it easy to directly compare to the internal representations of a model, such as the attention weights within a transformer (Fig. \ref{fig:fig6}). 
This makes it possible to track if the internal representations of a transformer computes the algebraic expression similarly to the algorithm encoded by its circuit -- the normative algorithm to computing polynomials.
More broadly, the distance metric that computes the distance between attention weights and ground truth circuit edges (as shown in Fig. \ref{fig:fig6}) can be used as a regularization term to encourage transformers to learn normative circuit computations, providing algorithmic and step-by-step information to the model.
Overall, evaluating AI models on algebraic problems allows practitioners to adjudicate between theoretical models of circuit computation with modern AI computation.

\subsection*{Evaluating LLMs with algebraic circuits}

We have introduced algebraic circuit tasks within the context of training models from scratch.
While it is difficult to evaluate pretrained LLMs with precise levels of certainty due to the obscure nature of the pretraining data and optimization protocols, algebraic tasks can still be repurposed for LLMs.
Prior work has demonstrated that LLMs typically fail on mathematical problems which require abstract reasoning on variables \citep{Nezhurina_alice_2024,mccoy_embers_2023,wu_reasoning_2024}.
Thus, if an LLM calls specific tools (e.g., calculators) which make computing circuits with only field elements too simple (e.g., those illustrated in Fig. \ref{fig:fig1} and \ref{fig:fig2}, we can replace these with circuits containing variables $x \in X$ (i.e., those of degree $\geq 1$)).
For example, such tasks can include the aforementioned polynomial identity testing or polynomial expansion/factorization tasks, both of which can be made arbitrarily difficult and use arbitrary sets of variables/tokens.

\begin{figure}[h!]
\centering
\includegraphics[width=3.43in]{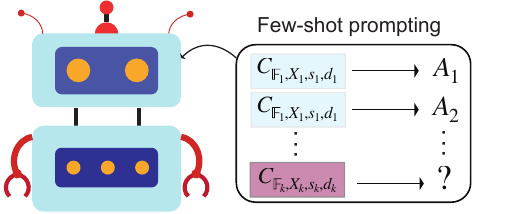}
\caption{The algebraic generalization capability across circuit divergence metrics can be evaluated through few-shot prompting in LLMs.
Given a set of question (left) and answer (right, $A_i$) pairs sampled from a specific circuit class $C_{\mathbb{F}_1,X_1,s_1,d_1}$ as prompts, we can ask an LLM to what degree it can generalize to algebraic problems sampled from a different circuit class $C_{\mathbb{F}_k,X_k,s_k,d_k}$.
}
\label{fig:fig7}
\end{figure}

Given that prior studies have demonstrated poor abstract reasoning abilities of LLMs, it is unlikely that current LLMs will be able to compute arbitrary circuits out-of-the-box.
However, recent techniques in prompting LLMs have suggested the ability of LLMs to learn from a few prompts (termed in-context learning) \citep{chan_transformers_2022,zhou_teaching_2023,reddy_mechanistic_2024,olsson_-context_2022,zhou_least--most_2022,wei_chain--thought_2022,nye_show_2021}.
In other words, a few question-answer pairs can be shown to the LLM in the context window, followed by the target problem.
This approach naturally offers an approach to measure algorithmic generalization across circuit divergence metrics: Given a class of question-answer pairs generated from a specific circuit class, how well can a model generalize to a problem generated from a different complexity class (Fig. \ref{fig:fig7})?
Another approach to systematically evaluate the ability of LLMs to compute circuits is through chain-of-thought prompting. 
Recent theoretical studies have indicated that leveraging chain-of-thought enables LLMs to solve more challenging problems by enhancing their expressivity and using scratchpads \citep{feng_towards_2023,strobl_what_2024,merrill_expressive_2024,li_chain_2024}.
Practically, inducing chain-of-thought in LLMs provides a way to verify whether LLMs sequentially ``reason'' through steps that are analogous to ground truth circuit computations (i.e., following the edges from input gates).
Moreover, since algebraic circuits encode an explicit algorithm to compute an expression, each step through the circuit produces a verifiable intermediate computation. 
This enables the evaluation of LLMs not only on verified outputs, but also verified intermediate reasoning chains, providing an opportunity to evaluate reinforcement learning approaches that have been increasingly used to develop reasoning-based models \citep{shao_deepseekmath_2024,deepseek-ai_deepseek-r1_2025}
Together, these experimental approaches provide concrete methods from which to quantify algorithmic ability in LLMs.

\section*{Open theoretical and empirical challenges}

Circuits provide a useful formalism to study algorithmic generalization in AI systems.
They are also a leading approach to quantifying the computational complexity of algorithmic problems in theoretical computer science.
Prior AI reasoning literature has focused on benchmarking AI to human-level reasoning\citep{mitchell_comparing_2023,legris_h-arc_2024,mcclelland_letting_2010,chollet_measure_2019}, or on how humans learn algorithmic tasks\citep{fedor_semantics_2012,lampinen_language_2024,lake_human-like_2023}, both of which are important areas of investigation from the perspectives of cognitive science and AI.
However, the present Perspective provides a complementary viewpoint from theoretical computer science aimed to quantify reasoning based on generalizing over algorithmic complexity.
Here we highlight the primary challenges associated with linking formal circuit models of computation with AI generalization.

\subsection*{Circuit complexity and AI generalization}
Algebraic circuit complexity studies the algorithmic resources required to evaluate polynomials. 
Though there are other metrics of complexity, such as Kolmogorov Complexity in algorithmic information theory \citep{kolmogorov_three_1968,li_introduction_2019}, this measure of complexity is incomputable, as it requires searching over an infinite number of programs (though there are efforts in approximating Kolmogorov Complexity with alternative methods; \citet{wyeth_circuit_2023,johnston_symmetry_2022,dingle_inputoutput_2018}).
Thus, while circuits allow for the explicit computation of a problem, it remains unclear as to whether notions from circuit complexity will naturally map onto notions of AI generalization.

Other studies have investigated the theoretical requirements for transformer models to evaluate formal languages and algorithms\citep{merrill_expressive_2024,strobl_what_2024,merrill_little_2025,yang_masked_2024,amiri_lower_2025}. 
While these studies provide useful insight into what architectural components are likely important to implement algorithmic problems of a particular complexity (e.g., deeper networks for circuit depth and wider context windows for circuit size), these studies do not address the learnability of circuit algorithms.
Furthermore, it will be interesting to understand how frameworks grounded in algorithmic circuit complexity (as presented here) relate to other measures of compositionality, which are often measured by properties of the computation graph of a task\citep{dziri_faith_2023}, divergence of input-output mapping of a task\citep{ram_what_2024}, or divergence of a distribution of training and testing tasks (e.g., MCD)\citep{keysers_measuring_2020}.

Nevertheless, given the lack of {\em any} framework to measure {\em algorithmic complexity} in AI, we believe that introducing a circuit complexity framework from which to design quantitative benchmarks will be an important step towards building a science of algorithmic generalization.
Furthermore, an algebraic circuit approach offers extensive machinery to enable generalization through other algebraic tools, such as minimum spanning/basis sets, their decompositions, and more.

\subsection*{Faithful algorithmic representation learning}

Though an algebraic circuit encodes an explicit algorithm to compute a polynomial at a particular level of abstraction (i.e., follow the edges from the input gates), it is possible that there are alternative viable algorithms to compute that same polynomial.
For example, would an AI model simply follow the edges from the input gates to the output gates? 
Or might it factorize the expression (leading to a shallower circuit) prior to evaluating that expression?
Relatedly, the algorithmic steps required to implement long addition (particularly when computing the sum of two very large numbers) are not fully captured in the algebraic circuits we present in this Perspective.
(For example, long addition as specified in \citet{dziri_faith_2023} requires additional operators such as ``carry'', ``concatenate'', ``modulo'', etc.)
Would AI models implement long addition with a different set of computational gates and operators to accommodate arithmetic with large numbers?
Nevertheless, despite these potential ambiguities, using a computational circuit framework provides \textit{testable} and \textit{verifiable} hypotheses that allow us to empirically evaluate what algorithm a model implements.
Furthermore, use of a circuit framework enables the design of quantitatively meaningful algorithmic benchmarks such as those designed to test generalization over algorithmic time complexity (circuit depth), or space complexity (circuit size), among others.
More broadly, a computational circuit framework can naturally extend beyond algebraic circuits to formal languages, such as those within the Chomsky hierarchy\citep{chomsky_three_1956} (e.g., context-free grammars), by characterizing the circuit parameters of the language's parse tree.
Thus, this framework provides a unified theoretical foundation for quantifying algorithmic complexity across diverse empirical evaluations, including arithmetic tasks, formal languages, and other compositional paradigms, and is an important step toward understanding the faithfulness by which an AI system computes a class of problems.

Alternatively, many prior approaches to faithfully learning algorithmic representations often involve neurosymbolic methods.
These methods provide promising avenues to learn discrete algorithmic solutions to problems that are reliable and sample efficient\citep{klinger_compositional_2023,lake_human-level_2015,poesia_peano_2023,trinh_solving_2024,ellis_dreamcoder_2021}.
However, designing general purpose (rather than domain-specific) neurosymbolic models can be challenging, since they are often not fully differentiable or require strong inductive biases.
By contrast, though statistical machine learning models (e.g., transformers) are ``general purpose'', the learning process is often obfuscated by learning dynamics that depend on architecture and complex optimization protocols.
This makes it difficult to ascertain what algorithms statistical systems learn.
However, recent studies in compositional representation learning have suggested that different factors -- such as choice of initialization and/or training curriculum -- can have a strong influence on whether a model learns compositionally \citep{lippl_when_2024,ito_compositional_2022,lake_human-like_2023,zhang_initialization_2024,saglietti_analytical_2022}.
In addition, developing techniques from the field of mechanistic interpretability provide new avenues from which to inspect whether the learned representations in an AI model are faithful to the hypothesized underlying algorithm (e.g., Fig. \ref{fig:fig6}; \citet{olsson_-context_2022,friedman_learning_2023}).
Nevertheless, leveraging diverse methods to carefully investigate the algorithms that symbolic and statistical AI models learn will be important for their interpretability, reliability, and overall safety to ensure reliable deployment of AI systems.

\section*{Conclusion}
Quantifying the algorithmic ability of AI systems is difficult due to the lack of a theoretical framework from which to establish meaningful benchmarks.
While there has been an increasing number of studies that have employed algebraic and compositional tasks to reliably elicit failure modes of transformers and LLMs, no theoretical framework exists to interpret these findings.
In this Perspective, we provide a parsimonious framework -- algebraic circuit complexity -- to evaluate the extent of a model's algorithmic generalization ability in terms of their circuit divergences.
In contrast to other formulations of complexity, such as Kolmogorov Complexity, encoding algorithmic problems as circuits provides an explicitly computable formulation.
The rich expressivity of algebraic problems, the data-rich nature of producing algebraic datasets, and the close links with algebraic circuits to other mathematical fields, makes algebraic circuit complexity a fruitful approach to quantify algorithmic generalization in AI systems.
More generally, the circuit complexity framework introduced here (with algebraic functions) can be naturally extended to other computational problems (e.g., formal languages, Boolean circuits).
We hope this Perspective provides the theoretical groundwork for future studies to quantify algorithmic generalization in modern AI systems with circuits.

\section*{Acknowledgments}

We thank Marco Carmosino and Karan Srivastava for helpful discussions on earlier versions of the manuscript. 
We acknowledge funding support from the Exploratory Science Councils at IBM Research.

\section*{Competing interests statement}
The authors declare no competing interests.

\bibliography{mybib}

\end{document}